\documentclass{article}
\usepackage{epsf}
\usepackage{latexsym}
\usepackage{comment}
\usepackage{times}
\usepackage[utf8x]{inputenc}
\usepackage[T1]{fontenc}
\usepackage{graphicx}
\usepackage{comment}
\usepackage{textcomp}
\usepackage{float}
\usepackage{subfigure}
\usepackage{setspace}
\usepackage{comment}
\usepackage{hyperref}
\usepackage{amssymb}
\usepackage{setspace}
\usepackage{amsmath}

\begin{document}

\title{Label-Dependencies Aware Recurrent Neural Networks}

\author{Yoann Dupont, Marco Dinarelli, Isabelle Tellier\\[18pt]
\small LaTTiCe (UMR 8094), CNRS, ENS Paris, Universit\'e Sorbonne Nouvelle - Paris 3\\
\small PSL Research University, USPC (Universit\'e Sorbonne Paris Cit\'e)\\
\small 1 rue Maurice Arnoux, 92120 Montrouge, France\\
\small yoa.dupont@gmail.com, ~ marco.dinarelli@ens.fr, ~ isabelle.tellier@univ-paris3.fr\\[30pt]
AUTHORS' DRAFT}

\maketitle

\begin{abstract}
In the last few years, Recurrent Neural Networks (RNNs) have proved effective on several NLP tasks.
Despite such great success, their ability to model \emph{sequence labeling} is still limited.
This lead research toward solutions where RNNs are combined with models which already proved effective in this domain, such as CRFs.
In this work we propose a solution far simpler but very effective: an evolution of the simple Jordan RNN, where labels are re-injected as input into the network, and converted into embeddings, in the same way as words.
We compare this RNN variant to all the other RNN models, Elman and Jordan RNN, LSTM and GRU, on two well-known tasks of Spoken Language Understanding (SLU).
Thanks to label embeddings and their combination at the hidden layer, the proposed variant, which uses more parameters than Elman and Jordan RNNs, but far fewer than LSTM and GRU, is more effective than other RNNs, but also outperforms sophisticated CRF models.
\end{abstract}

\section{Introduction}
\label{sec:Intro}

In the last few years Recurrent Neural Networks (RNNs) \cite{jordan-serial,Elman90findingstructure,Hochreiter-1997-LSTM} have proved very effective in several Natural Language Processing (NLP) tasks such as Part-of-Speech tagging (POS tagging), chunking, Named Entity Recognition (NER), Spoken Language Understanding (SLU), machine translation and even more \cite{RNN-Mikolov-Interspeech-2010,RNNExtensions_Mikolov-ICASSP-2011,Collobert-2008-UAN-1390156.1390177,Collobert-2011-NLP-1953048.2078186,RNNforLU-Interspeech-2013,RNNforSLU-Interspeech-2013,Vukotic.etal_2015}.
These models are particularly effective thanks to their recurrent architecture, which allows neural models to \textit{keep in memory} past information and re-use it at the current processing step.

In the literature of RNNs applied to NLP, several architectures have been proposed.
At first Elman and Jordan RNNs, introduced in \cite{Elman90findingstructure,jordan-serial}, and known also as simple RNNs, have been adapted to NLP.
The difference between these two models is in the type of connection giving the recurrent character to these two architectures: in the Elman RNN the recursion is a loop at the hidden layer, while in the Jordan RNN it relies the output layer to the hidden layer.
This last recursion allows to use at the current step labels predicted for previous positions in a sequence.

These two recurrent models have shown limitations in learning relatively long contexts \cite{Bengio-1994-RNN-Learning-Difficulty}.
In order to overcome this limitation the RNNs known as \textit{Long Short-Term Memory} (LSTM) have been proposed \cite{Hochreiter-1997-LSTM}.
Recently, a simplified and, apparently, more effective variant of LSTM has been proposed, using \textit{Gated Recurrent Units} and thus named \textit{GRU} \cite{Cho-2014-GatedRecurrentUnits}.

Despite outstanding performances on several NLP tasks, RNNs have not been explicitly adapted to integrate effectively label-dependency information in sequence labeling tasks.
Their sequence labeling decisions are based on intrinsically local functions (e.g. the softmax).
In order to overcome this limitation, sophisticated hybrid \textit{RNN+CRF} models have been proposed \cite{huang2015bidirectional,lample2016neural,Ma-Hovy-ACL-2016}, 
where the traditional output layer is replaced by a CRF neural layer.
These models reach state-of-the-art performances, their evaluation however is not clear.
In particular it is not clear if performances derive from the model itself, or thanks to particular experimental conditions.
In \cite{Ma-Hovy-ACL-2016} for example, the best result of POS tagging on the Penn Treebank corpus is an accuracy of $97.55$,  which is reached using word embeddings trained using \textit{GloVe} \cite{pennington2014glove}, 
on huge amount of unlabeled data. The model of \cite{Ma-Hovy-ACL-2016} without pre-trained embeddings reaches an accuracy of $96.9$, which doesn't seem that outstanding if we consider that a CRF model dating from $2010$, trained from scratch, without using any external resource, reaches an accuracy of $97.3$ on the same data \cite{lavergne2010practical}. We achieved the same result on the same data with a CRF model trained from scratch using the incremental procedure described in \cite{Dinarelli:Rosset:ijcnlp:2011}.
Moreover, the first version of the network proposed in this paper, but using a sigmoid activation function and only the $L_2$ regularization, tough with a slightly different data preprocessing, achieves an accuracy on the Penn Treebank of $96.9$ \cite{2016:arXiv:DinarelliTellier:NewRNN}.

The intuition behind this paper is that embeddings allow a fine and effective modeling not only of words, 
but also of labels and label dependencies, which are crucial in some tasks of sequence labeling.
In this paper we propose, as alternative to \textit{RNN+CRF} models, a variant of RNN allowing this more effective modeling.
Surprisingly, a simple modification to the RNN architecture results in a very effective model: 
in our variant of RNN the recurrent connection connects the output layer to the input layer and, 
since the first layer is just a \textit{look-up} table mapping discrete items into embeddings, 
labels predicted at the output layer are mapped into embeddings the same way as words.
Label embeddings and word embeddings are combined at the hidden layer, allowing to learn relations between these two types of information, which are used to predict the label at current position in a sequence.
Our intuition is that using several label embeddings as context, a RNN is able to model correctly label-dependencies, the same way as more sophisticated models explicitly designed for sequence labeling like CRFs \cite{lafferty01:crf}.

This paper is a straight follow-up of \cite{DinarelliTellier:RNN:CICling2016}.
Contributions with respect to that work are as follows:

i) An analysis of performances of \textit{forward}, \textit{backward} and bidirectional models.
ii) The use of \textit{ReLU} hidden layer and \textit{dropout} regularization \cite{JMLR-v15-srivastava14a} at the hidden and embedding layers for improved regularized models.
iii) The integration of a character-level convolution layer.
iv) An in-depth evaluation, showing the effect of different components and of different information level on the performance.
v) A straightforward comparison of the proposed variant of RNN to Elman, Jordan, LSTM and GRU RNNs, 
showing that the new variant is at least as effective as the best RNN models, such as LSTM and GRU.
Our variant is even more effective when taking label-dependencies into account is crucial in the task, proving that our intuition is correct.

\begin{figure}%[width=\linewidth]%[t, width=\linewidth]
  \centering
        \footnotesize
  \subfigure[Elman]{
    \includegraphics[height=4.2cm]{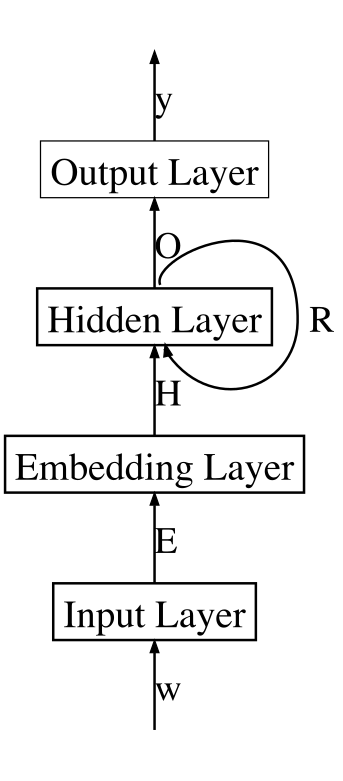}
  }
  \quad
  \subfigure[Jordan]{
    \includegraphics[height=4.2cm]{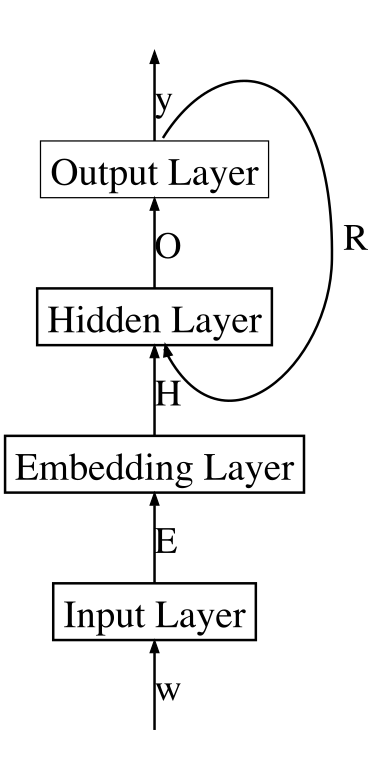}
  }
  \quad
  \subfigure[Our variant]{
    \includegraphics[height=4.2cm]{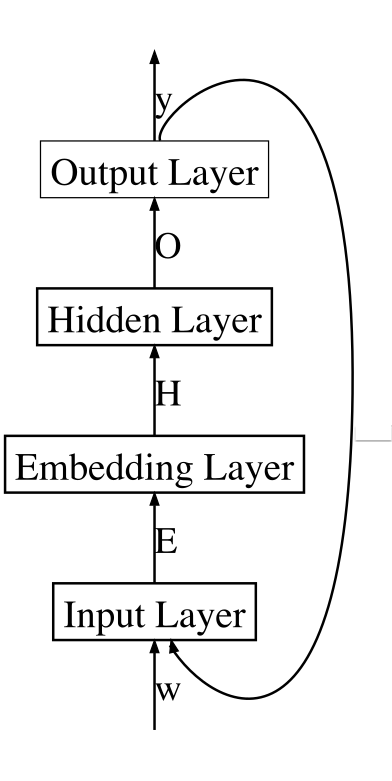}
  }
  \vspace{-0.8em}
  \caption{High level schema of simple RNNs (Elman and Jordan) and the variant proposed in this paper.}
	\label{fig:3architecgtures}
        \vspace{-1.5em}
\end{figure}

An high level schema of simple RNNs and of the variant proposed in this paper is shown in figure~\ref{fig:3architecgtures}, where $w$ is the input word, $y$ is the label, $E$, $H$, $O$ and $R$ are the model parameters, which will be discussed in the following sections.

Since evaluations on tasks like POS tagging on the Penn Treebank are basically reaching perfection (state-of-the-art is at $97.55$ accuracy), any new model would probably provide little or no improvement.
Also, performances on this type of tasks seem to have reached a \textit{plateau}, as models achieving $97.2$ accuracy or even better, were already published starting from $2003$ \cite{Toutanova03posTagging,shen-satta-joshi:2007:ACLMain}.
We propose instead to evaluate all the models on two different and widely used tasks of Spoken Language Understanding \cite{demori08-SPM}, which provide more variate evaluation settings: ATIS \cite{Dahl-1994-ESA-1075812.1075823} and MEDIA \cite{Bonneau-Maynard2006-media}.

ATIS is a relatively simple task and doesn't require a sophisticated modeling of label dependencies.
This task allows to evaluate models in similar settings as tasks like POS tagging or Named Entity Recognition as defined in the \textit{CoNLL Shared Task} 2003, both widely used as benchmarks in NLP papers.
MEDIA is a very challenging task, where the ability of models to keep label dependencies into account is crucial to obtain good results.

Results show that our new variant is as effective as the best RNN models on a simple task like ATIS, still having the advantage of being much simpler.
On the MEDIA task however, our variant outperforms all the other RNNs by a large margin, and even sophisticated CRF models, providing the best absolute result ever achieved on this task.

The paper is organized as follows:
In the next section we describe the RNNs used in the literature for NLP, starting from existing models to arrive at describing the new variant we propose.
In the section~\ref{sec:Eval} we present the corpora used for evaluation, the experimental settings and the results obtained in several experimental conditions. 
We draw some conclusions in section~\ref{sec:Conclusions}.

\vspace{-1.5em}

\section{Recurrent Neural Networks (RNNs)}
\label{sec:RNN}

In this section we describe the most popular RNNs used for NLP, such as Elman and Jordan RNNs \cite{jordan-serial,Elman90findingstructure}, and the most sophisticated RNNs like LSTM and GRU \cite{Hochreiter-1997-LSTM,Cho-2014-GatedRecurrentUnits}.
We also describe training and inference procedures, and the RNN variant we propose.
\vspace{-1.0em}

\subsection{Elman and Jordan RNNs}
\label{subsec:OldRNNs}

Elman and Jordan RNNs are defined as follows:

\vspace{-1.0em}
\begin{equation}
\label{eqn:ElmanHidden}
\mathbf{h_t}^{\text{Elman}} = \Phi(R ~ \mathbf{h_{t-1}^{\text{Elman}}} + H ~ \mathbf{I_t})
\end{equation}

\vspace{-2.0em}
\begin{equation}
\label{eqn:JordanHidden}
\mathbf{h_t}^{\text{Jordan}} = \Phi(R ~ \mathbf{y_{t-1}} + H ~ \mathbf{I_t})
\end{equation}
\vspace{-1.0em}

\noindent The difference between these two models is in the way of computing hidden activities, 
while the output is computed in the same way:

\vspace{-1.2em}
\begin{equation}
\mathbf{y_t} = softmax(O ~ \mathbf{h_t^{*}})
\end{equation}
\vspace{-1.5em}

\noindent $h_t^{*}$ and $y_t$ are respectively the hidden and output layer's activities\footnote{$h_*$ means the hidden layer of any model, as the output layer is computed in the same way for all networks described in this paper.}, $\Phi$ is an activation function, 
$H$, $O$ and $R$ are the parameters at the hidden, output and recurrent layer, respectively (biases are omitted to keep equations lighter).
$h_{t-1}^{\text{Elman}}$ is the hidden layer activity computed at previous time step and used as context in the Elman RNN, 
while $y_{t-1}$ is the previous predicted labels, used as context in the Jordan RNN.
$I_t$ is the input, which is often the concatenation of word embeddings in a fixed window $d_w$ (for win\textbf{D}ow of \textbf{W}ords) around the current word $w_t$ to be labeled.
We define as $E(w_i)$ the embedding of any word $w_i$.
$I_t$ is then defined as:

\vspace{-1.3em}
\begin{equation}
\mathbf{I_t} = [E_w(w_{t-d_w}) ... E_w(w_{t}) ... E_w(w_{t+d_w})]
\end{equation}
\vspace{-1.5em}

\noindent where $[ ~ ]$ is the concatenation of vectors (or matrices in the following sections).
The $softmax$ function, given a set $S$ of $m$ numerical values $v_i$, 
associated to discrete elements $i \in [1,m]$, computes the probability associated to each element as:

{\centering
$\forall i \in [1,m]$~$p(i) = \frac{e^{v_i}}{\sum_{j=1}^m e^{v_j}}$

}

\noindent This function allows to compute the probability associated to each label and choose as predicted label the one with the highest probability.
\vspace{-0.8em}

\subsection{\textit{Long Short-Term Memory} (LSTM) RNNs}
\label{subsec:LSTM}

While LSTM is often used as the name of the whole network, it just defines a different way of computing the hidden layer activities.
LSTMs use \textit{gate} units to control how past and present information affect the network's internal state, and a \textit{cell} to store past information that is going to be used as context at the current processing step.
 \textit{Forget}, \textit{input} gates and \textit{cell state} are computed as:

\vspace{-1.8em}
\begin{eqnarray}
\mathbf{f_t} & = & \Phi( W_f \mathbf{h_{t-1}} + U_f \mathbf{I_t} ) \\
\mathbf{i_t} & = & \Phi( W_i \mathbf{h_{t-1}} + U_i \mathbf{I_t} ) \\
\mathbf{\hat{c}_t} & = & \Gamma( W_c \mathbf{h_{t-1}} + U_c \mathbf{I_t} )
\end{eqnarray}
\vspace{-1.5em}

\noindent $\Gamma$ is used to indicate a different activation function from $\Phi$\footnote{In the literature $\Phi$ and $\Gamma$ are the \textit{sigmoid} and \textit{tanh}, respectively}.
$\hat{c}_t$ is actually an intermediate value used to update the \textit{cell state} value as follows:

\vspace{-1.2em}
\begin{equation}
\mathbf{c_t = f_t \odot c_{t-1} + i_t \odot \hat{c_t}}
\end{equation}
\vspace{-1.5em}

\noindent $\odot$ is the element-wise multiplication. Once these quantities have been computed, the \textit{output} gate is computed and used to control the hidden layer activities at the current time step $t$:

\vspace{-1.8em}
\begin{eqnarray}
\mathbf{o_t} & = & \Phi( W_o \mathbf{h_{t-1}} + U_o \mathbf{I_t} ) \\
\mathbf{h_t}^{\text{LSTM}} & = & \mathbf{o_t} \odot \Phi( \mathbf{c_t} )
\end{eqnarray}
\vspace{-1.5em}

\noindent Once again (and in the remainder of the paper), biases are omitted to keep equations lighter.
As we can see, each gate and the cell state have their own parameter matrices $W$ and $U$, 
used for the linear transformation of the previous hidden state ($h_{t-1}$) and the current input ($I_t$).
The evolution of the LSTM layer named GRU (\textit{Gated Recurrent Units}) \cite{Cho-2014-GatedRecurrentUnits}, 
combines together \textit{forget} and \textit{input} gates, and the previous hidden layer with the \textit{cell state}:

\vspace{-1.8em}
\begin{eqnarray}
\mathbf{z_t} & = & \Phi( W_z \mathbf{h_{t-1}} + U_z \mathbf{I_t} ) \\
\mathbf{r_t} & = & \Phi( W_r \mathbf{h_{t-1}} + U_r \mathbf{I_t} ) \\
\mathbf{\hat{h_t}} & = & \Gamma( W (\mathbf{r_t \odot  h_{t-1}}) + U \mathbf{I_t} ) \\
\mathbf{h_t}^{\text{GRU}} & = & (1 - \mathbf{z_t}) \odot \mathbf{h_{t-1}} + \mathbf{z_t \odot \hat{h_t}}
\end{eqnarray}
\vspace{-1.5em}

\noindent GRU is thus a simplification of LSTM, it uses less units and it has less parameters to learn.

\subsection{LD-RNN : Label-Dependencies Aware Recurrent Neural Networks}
\label{subsec:I-RNN}

The variant of RNN that we propose in this paper can be thought of as having a recurrent connection from the output to the input layer. Note that from a different perspective, this variant can just be seen as a Feed-Forward Neural Network (FFNN) using previous predicted labels as input. Since Jordan RNN has the same architecture, the only difference being that in contrast to Jordan models we embed labels, we still prefer talking about recurrent network.
This simple modification to the architecture of the network has important consequences on the model.

The reason motivating this modification is that we want embeddings for labels and use them the same way as word embeddings.
Like we mentioned in the introduction, the first layer is a look-up table mapping discrete, or \textit{one-hot}\footnote{The one-hot representation of a token represented by an index $i$ in a dictionary, is a vector $v$ of the same size as the dictionary and assigned zero everywhere, except at position $i$ where it is $1$.}, representations into distributional representations.

Such representations can encode very fine syntactic and semantic properties, as it has already been proved by \textit{word2vec} \cite{Word2Vec_Mikolov-2013} or \textit{GloVe} \cite{pennington2014glove}.
We want similar properties to be learned also for labels, so that to encode in label embeddings the label dependencies needed for sequence labeling tasks.
In this paper we learn label embeddings from the sequences of labels associated to word sentences in annotated data.
But this procedure could be applied also when structured label information is available.
We could thus exploit syntactic parse trees, structured named entities or entity relations for learning sophisticated label embeddings.

The idea of using label embeddings has been introduced in \cite{chen-manning-2014-EMNLP2014} for dependency parsing, resulting in a very effective parser.
In this paper we go ahead with respect to \cite{chen-manning-2014-EMNLP2014} by using several label embeddings as context to predict the label at current position in a sequence. Also we pre-train label embeddings like it is usually done for words. 
As consequence, we learn first generic dependencies between labels without their interactions with words. Such interactions are then integrated and refined during the learning phase of the target sequence labeling task.
For this ability to learn label-dependencies, we name our variant \emph{LD-RNN}, 
standing for \textit{Label Dependencies aware RNN}.

Using the same formalism as before, 
we define $E_w$ the matrix for word embeddings, 
while $E_l$ is the matrix for label embeddings.
The word-level input to our RNN is $I_t$ as for the other RNNs, while the label-level input is:

\vspace{-1.2em}
\begin{equation}
\mathbf{L_t} = [E_l(y_{t-d_l+1}) ~ E_l(y_{t-d_l+2}) \dots E_l(y_{t-1})]
\end{equation}
\vspace{-1.5em}

\noindent which is the concatenation of vectors representing the $d_l$ previous predicted labels ($d_l$ stands (for win\textbf{D}ow of \textbf{L}abels)).
The hidden layer activities of our RNN variant are computed as:

\vspace{-1.0em}
\begin{equation}
\mathbf{h_t}^{\text{LD-RNN}} = \Phi( H ~ [\mathbf{I_t L_t}] )
\end{equation}
\vspace{-1.5em}

\noindent We note that we could rewrite the equation above as $\Phi(H_w \mathbf{I_t} + H_l \mathbf{L_t})$ with a similar formalism as before, the two equations are equivalent if we define $H = [H_w H_l]$.

Thanks to the use of label embeddings and their combination at the hidden layer, 
our \emph{LD-RNN} variant learns very effectively label dependencies.
Since the other RNNs in general don't use explicitly the label information as context, they can predict incoherent label sequences.
As we already mentioned, this limitation lead research toward hybrid \emph{RNN+CRF} models \cite{huang2015bidirectional,lample2016neural,Ma-Hovy-ACL-2016}.

Another consequence of the modification introduced in our RNN variant is an improved robustness to prediction mistakes.
Since we use several label embeddings as context (see $\mathbf{L_t}$ above), 
once the model has learned label embeddings, in the test phase it is unlikely that several prediction mistakes occur in the same context. Even in that case, thanks to properties encoded in the embeddings, mistaken labels have similar representations to correct labels, allowing the model to possibly predict correct labels.
Reusing an example from \cite{LinguisticRegularities-Mikolov-2013}:
if \textit{Paris} is replaced by \textit{Rome} in a text, this has no impact on several NLP tasks, 
as they are both proper nouns in POS tagging, localization in Named Entity Recognition etc.
Using label embeddings provides the \emph{LD-RNN} variant with the same robustness on the label side.

While the traditional Jordan RNN uses also previous labels as context information, 
it has not the same robustness because of the poor label representation used in adaptations of this model to NLP tasks.
In Jordan RNNs used for NLP like \cite{RNNforLU-Interspeech-2013,RNNforSLU-Interspeech-2013,Vukotic.etal_2015},  labels are represented either with the probability distribution computed by the $softmax$, 
or with the \textit{one-hot} representation computed from the probability distribution.

In the latter case it is clear that a prediction mistake can have a bad impact in the context, 
as the only value being $1$ in the \textit{one-hot} representation would be in the wrong position.
Instead, using the probability distribution may seem a kind of \textit{fazzy} representation over several labels, 
but we have found empirically that the probability is very sharp and picked on one or just few labels.
In any case this representation doesn't provide the desired robustness that can be achieved with label embeddings.

From another point of view, we can interpret the computation of the hidden activities in a Jordan RNN as using label embeddings. In the equation \ref{eqn:JordanHidden}, the multiplication $R y_{t-1}$, since $y_{t-1}$ is a sparse vector, 
can be interpreted as the selection of an embedding from $R$.

Even with this interpretation there is a substantial difference between a Jordan RNN and our variant.
In the Jordan RNN, once the label embedding has been computed with $R y_{t-1}$, 
the result is not involved in the linear transformation applied by the matrix $H$, which is only applied to the word-level input $I_t$.
The result of this multiplication is added to $R y_{t-1}$ and then the activation function is applied.

In our variant in contrast, labels are first mapped into embeddings with $E[y_i]$\footnote{In our case, $y_i$ is explicitly converted from probability distribution to \textit{one-hot} representation.}.
Word and label inputs $\mathbf{I_t}$ and $\mathbf{L_t}$ are then both transformed by multiplying by $H$, 
which is correctly dimensioned to apply the linear transformation on both inputs.
In our variant thus, two different label transformations are always applied: i) the conversion from sparse to embedding representation; ii) the linear transformation by multiplying label embeddings by $H$.
\vspace{-1.0em}

\subsection{Learning and Inference}
\label{subsec:Learning}

We learn the \emph{LD-RNN} variant like all the other RNNs, by minimizing the cross-entropy between the expected label $l_t$ and the predicted label $y_t$ at position $t$ in the sequence, plus a $L_2$ regularization term:

\vspace{-1.2em}
\begin{equation}
C = - \mathit{l_t} \odot  log( \mathbf{y_t} ) + \frac{\lambda}{2} \left | \Theta \right |^2
\end{equation}
\vspace{-1.5em}

\noindent $\lambda$ is a hyper-parameter to be tuned, $\Theta$ is a short notation for $E_w, E_l, H, O$.
$l_t$ is the \textit{one-hot} representation of the expected label.
Since $y_t$ above is the probability distribution over the label set, 
we can see the output of the network as the probability $P(i | \mathbf{I_t, L_t})$~$\forall i \in [1,m]$, 
where $\mathbf{I_t}$ and $\mathbf{L_t}$ are the input of the network (words and labels), $i$ is the index of one of the labels defined in the targeted task.

We can thus associate to the \emph{LD-RNN} model the following decision function:

\vspace{-1.2em}
\begin{equation}
argmax_{i \in [1,m]} P(i | \mathbf{I_t, L_t})
\end{equation}
\vspace{-1.5em}

\noindent We note that this is still a local decision function, as the probability of each label is normalized at each position of a sequence.
Despite this, the use of label-embeddings $\mathbf{L_t}$ as context allows the \emph{LD-RNN} to effectively model label dependencies.
Since the other RNNs like Elman and LSTM don't use the label information in their context, 
their decision function can be defined as:

\vspace{-1.2em}
\begin{equation}
argmax_{i \in [1,m]} P(i | \mathbf{I_t})
\end{equation}
\vspace{-1.5em}

\noindent which can lead to incoherent predicted label sequences.

We use the traditional back-propagation algorithm with momentum to learn our networks \cite{PracticalRecommendations-Bengio-2012}.
Given the recurrent nature of the networks, the Back-Propagation Through Time (BPTT) is often used \cite{werbos-bptt}.
This algorithm consists in unfolding the RNN for $N$ previous steps, $N$ being a parameter to choose, and using thus the $N$ previous inputs and hidden states to update the model's parameters.
The traditional back-propagation algorithm is then applied. This is equivalent to learn a feed-froward network of  depth $N$.
The BPTT algorithm is supposed to allow the network to learn arbitrary long contexts.
However \cite{RNNExtensions_Mikolov-ICASSP-2011} has shown that RNNs for language modeling learn best with only $N = 5$ previous steps. 
This can be due to the fact that, at least in NLP, a longer context does not lead necessarily to better performances, 
as a longer context is also more noisy.

Since the BPTT algorithm is quite expensive, 
\cite{RNNforSLU-Interspeech-2013} chose to explicitly use the contextual information provided by the recurrent connection, and to use the traditional back-propagation algorithm, apparently without performance loss.

In this paper we use the same strategy.
When the contextual information is used explicitly in a Jordan RNN, the hidden layer state is computed as follows:

\vspace{-1.2em}
\begin{equation}
\mathbf{h_t} = \Phi(R [\mathbf{y_{t-d_l+1} \; y_{t-d_l+2}} \; ... \; \mathbf{y_{t-1}} ] + H ~ \mathbf{I_t})
\end{equation}
\vspace{-1.5em}

\noindent A similar modification can be applied also to Elman, LSTM and GRU RNNs to keep into account explicitly the previous hidden states. To our knowledge however, these networks are effectively learned using only one previous hidden state \cite{huang2015bidirectional,lample2016neural,Ma-Hovy-ACL-2016}.

From explanations above we can say that using explicit wide context of words and labels like we do in LD-RNN, can be seen as an approximation of the BPTT algorithm.

\vspace{-0.8em}

\subsection{Toward More Sophisticated Networks: Character-Level Convolution}
\label{subsec:Improvements}

%\subsubsection{Character-Level Convolution}
%\label{subsec:CNN}

Even if word embeddings provide a very fine encoding of word features, several works such like \cite{huang2015bidirectional,LSTM-CNN-NER-2015,lample2016neural,Ma-Hovy-ACL-2016} have shown that more effective models can be obtained using a convolution layer over characters of words.
Character-level information is indeed very useful to allow a model generalizing over rare inflected surface forms and even out-of-vocabulary words in the test phase. Word embeddings are in fact much less effective in such cases.
The convolution over word characters provide also the advantage of being very general: 
it can be applied in the same way to different languages, allowing to re-use the same system on different languages and tasks.

In this paper we focus on a convolution layer similar to the one used in \cite{Collobert-2011-NLP-1953048.2078186} for words.
For any word $w$ of length $|w|$, we define $E_{ch}(w,i)$ the embedding of the character $i$ of the word $w$.
We define $W_{ch}$ the matrix of parameters for the linear transformation applied by the convolution (once again we omit the associated bias).
We compute a convolution of window size $2 d_c + 1$ over characters of a word $w$ as follows:

\begin{itemize}
\item $\forall i \in [1,|w|]$
$Conv_i = W_{ch} [E_{ch}(w,i-d_c); \dots E_{ch}(w,i); \dots E_{ch}(w,i+d_c)]$

\item $Conv_{ch} = [Conv_1 \dots Conv_{|w|}]$

\item $Char_{w} = Max(Conv_{ch})$
\end{itemize}

\noindent the $Max$ function is the so-called max-pooling \cite{Collobert-2011-NLP-1953048.2078186}.
While it is not strictly necessary mapping characters into embeddings, it would be probably less interesting applying the convolution on discrete representations.
The matrix $Conv_{ch}$ is made of the concatenation of vectors returned from the application of the linear transformation $W_{ch}$. Its size is thus $|C| \times |w|$, where $|C|$ is the size of the convolution layer.
The max-pooling computes the maxima over the word-length direction, 
thus the final output $Char_{w}$ has size $|C|$, which is independent from the word length.
$Char_{w}$ can be interpreted as a distributional representation of the word $w$ encoding the information at $w$'s character level. This is a complementary information with respect to word embeddings, which encode inter-word information, and provide the model with an information similar to what is provided by discrete lexical features like word prefixes, suffixes, capitalization information etc., plus information about morphologically correct words of a given language.

\subsection{RNN Complexities}
\label{subsec:Complexity}

The improved modeling of label dependencies in our \emph{LD-RNN} variant is achieved at the cost of more parameters with respect to the simple RNN models.
However the number of parameters is still much less than sophisticated networks like LSTM.
In this section we provide a comparison of RNNs complexity in terms of the number of parameters.

We introduce the following symbols: $|H|$ and $|O|$ are the size of the hidden and output layers, respectively. The size of the output layer is the number of labels; $N$ is the embedding size, in \emph{LD-RNN} we use the same size for word and label embeddings; $d_w$ is the window size used for context words; and $d_l$ is the number of label embeddings we use as context in \emph{LD-RNN}.
We analyze the hidden layer of all networks, and the embedding layer for \emph{LD-RNN}.
The other layers are exactly the same for all the networks described in this paper.

For Elman and Jordan RNNs, the hidden layer has the following number of parameters, respectively:

{\centering
$\{|H| * |H|\}_R + \{|H| * (2 d_w + 1) N\}_{H^{\text{Elman}}}$

}

{\centering
$\{|O| * |H|\}_R + \{|H| * (2 d_w + 1) N\}_{H^{\text{Jordan}}}$

}

\noindent Subscripts indicate from which matrix the parameters come.
The factor $(2 d_w + 1) N$ comes from the $(2 d_w + 1)$ words used as input context and then mapped into embeddings.
The factor $|O| * |H|$ in Jordan RNN is due to the fact that the matrix $R$ connects output and hidden layers.

In \emph{LD-RNN} we have:

{\centering
$\{|O| * N\}_{E_l} + \{((2 d_w + 1 + d_l) N) * |H|\}_{H^{\text{LD-RNN}}}$

}

The factor $|O| * N$ is due to the use of the matrix $E_l$ containing $|O|$ label embeddings of size $N$.
Since in this paper we chose $N = |H|$ and $|O| < |H|$, and since in LD-RNN we don't use any matrix R on the recurrent connection, the fact of using label embeddings doesn't increase the number of parameters of the \emph{LD-RNN} variant.

The hidden layer of \emph{LD-RNN} however is dimensioned to connect all the word and label embeddings to all the hidden neurons. As consequence in the matrix $H$ we have $d_l N$ more parameters than in the matrix $H$ of Elman and Jordan RNNs.

In LSTM and GRU RNNs we have two extra matrices $W$ and $U$ for each gate and for the \textit{cell state}, 
used to connect the previous hidden layer and the current input, respectively.
These two matrices contain thus $|H| * |H|$ and $(2 wd + 1) N * |H|$ parameters, respectively.

Using the same notation and the same settings as above, in the hidden layer of LSTM and GRU we have the following number of parameters:

{\centering
$\{4 (|H| * |H| + |U| * (2 d_w + 1) N)\}_{H^{\text{LSTM}}}$

}

{\centering
$\{3 (|H| * |H| + |U| * (2 d_w + 1) N)\}_{H^{\text{GRU}}}$

}

The $3$ for GRU reflects the fact that this network uses only $2$ gates and a \textit{cell state}.
It should be pointed out, however, that while we have been testing LSTM and GRU with a word window for a matter of fair comparison\footnote{Indeed we observed better performances when using a word window with respect to when using a single word}, these layers are applied on the current word and the previous hidden layer only, without the need of a word window. This is because this layer learns automatically how to use previous word information.
In such case the complexity of the LSTM layer reduces to $\{4 (|H| * |H| + |U| * N)\}_{H^{\text{LSTM}}}$.
If we choose $|U| = |H|$, such complexity is comparable to that of \emph{LD-RNN} in terms of number of parameters (slightly less actually). The LSTM is still more complex however because the hidden layer computation requires $4$ gates and the cell state ($\mathbf{\hat{c_t}}$) computations (each involving $2$ matrix multiplications), the update of the new cell state $\mathbf{c_t}$ (involving also $2$ matrix multiplications), and only after the hidden state can be computed.
\emph{LD-RNN}'s hidden state, in contrast, requires only matrix rows selection and concatenation to compute $\mathbf{I_t}$ and $\mathbf{L_t}$, which are very efficient operations, and then the hidden state can already be computed. 

As consequence, while the variant of RNN we propose in this paper is more complex than simple RNNs, 
LSTM and GRU RNNs are by far the most complex networks.

\subsection{\textit{Forward}, \textit{Backward} and Bidirectional Networks}
\label{subsec:bidir}

The RNNs introduced in this paper are proposed as forward, backward and bidirectional models \cite{Schuster-1997-BRNN}.
The forward model is what has been described so far.
The architecture of the backward model is exactly the same, 
the only difference is that the backward model processes data from the end to the begin of sequences. Labels and hidden layers computed by the backward model can thus be used as future context in a bidirectional model.

Bidirectional models are described in details in \cite{Schuster-1997-BRNN}.
In this paper we utilize the version using separate forward and backward models.
The final output is computed as the geometric mean of the output of the two individual models, that is:

{\centering
$\mathbf{y_t} = \sqrt{\mathbf{y_t^f} \odot \mathbf{y_t^b}}$

}

where $\mathbf{y_t^f}$ and $\mathbf{y_t^b}$ are the output of the forward and backward models, respectively.

In the development phase of our systems, we noticed no difference in terms of performance between the two types of bidirectional models described in \cite{Schuster-1997-BRNN}. We chose thus the version described above, since it allows to initialize all the parameters with the forward and backward models previously trained.
As consequence the bidirectional model is very close to a very good optimum since the first learning iteration, and very few iterations are needed to learn the final model.

\section{Evaluation}
\label{sec:Eval}

\subsection{Corpora for Spoken Language Understanding}
\label{subsec:Corpora}

We evaluated our models on two tasks of Spoken Language Understanding (SLU) \cite{demori08-SPM}:

\textbf{The ATIS corpus} (\textit{Air Travel Information System}) \cite{Dahl-1994-ESA-1075812.1075823} was collected for building a spoken dialog system able to provide flight information in the United States.

ATIS is a simple task dating from $1993$.
Training data are made of $4978$ sentences chosen among dependency-free sentences in the \texttt{ATIS-2} and \texttt{ATIS-3} corpora.
The test set is made of $893$ sentences taken from the \texttt{ATIS-3} \texttt{NOV93} and \texttt{DEC94} data.
Since there are not official development data, we taken a part of the training set for this purpose.
The word and label dictionaries contain $1117$ and $85$ items, respectively.
We use the version of the corpus published in \cite{raymond07-luna}, 
where some word classes are available, such as city names, airport names, time expressions etc.
These classes can be used as features to improve the generalization of the model on rare or unseen words.
More details about this corpus can be found in \cite{Dahl-1994-ESA-1075812.1075823}.

An example of utterance transcription taken from this corpus is \textit{``I want all the flights from Boston to Philadelphia today''}.
The words \textit{Boston}, \textit{Philadelphia} and \textit{today} in the transcription are associated to the concepts \emph{DEPARTURE.CITY}, \emph{ARRIVAL.CITY} and \emph{DEPARTURE.DATE}, respectively.
All the other words don't belong to any concept, they are associated to the void concept named \emph{O} (for Outside).
This example show the simplicity of this task:
the annotation is sparse, only $3$ words of the transcription are associated to a non-void concept;
there is no segmentation problem, as each concept is associate to one word.
Because of these two characteristics, the ATIS task is similar on the one hand to a POS tagging task, where there is no segmentation of labels over multiple words; on the other hand it is similar to a linear Named Entity Recognition task, where the annotation is sparse.

We are aware of the existence of two version of the ATIS corpus:
the official version published starting from \cite{raymond07-luna}, 
and the version associated to the tutorial of \emph{deep learning} made available by the authors of \cite{RNNforSLU-Interspeech-2013}.\footnote{Available at http://deeplearning.net/tutorial/rnnslu.html}.
This last version has been modified, some proper nouns have been re-segmented (for example the token \textit{New-York} has been replaced by two tokens \textit{New York}), and a preprocessing has been applied to reduce the word dictionary (numbers have been converted into the conventional token \emph{DIGIT}, and singletons of the training data, as well as out-of-vocabulary words of the developpement and test data, have been converted into the token \emph{UNK}).
Following the tutorial of \cite{RNNforSLU-Interspeech-2013} we have been able to download the second version of the ATIS corpus. However in this version word classes that are available in the first version are not given.
We ran some experiments with these data, using only words as input. The results we obtained are comparable with those published in \cite{Mesnil-RNN-2015}, in part from same authors of \cite{RNNforSLU-Interspeech-2013}.
However without word classes we cannot fairly compare with works that are using them.
In this paper we thus compare only with published works that used the official version of ATIS.

\textbf{The French corpus MEDIA} \cite{Bonneau-Maynard2006-media} was collected to create and evaluate spoken dialog systems providing touristic information about hotels in France.
This corpus is made of $1250$ dialogs collected with \textit{Wizard-of-OZ} approach.
The dialogs have been manually transcribed and annotated following a rich concept ontology.
Simple semantic components can be combined to create complex semantic structures.\footnote{For example the component \emph{localization} can be combined with other components like \texttt{city}, \texttt{relative-distance}, \texttt{generic-relative-location}, \texttt{street} etc.}
The rich semantic annotation is a source of difficulties, but also the annotation of coreference phenomena.
Some words cannot be correctly annotated without knowing a relatively long context, 
often going beyond a single dialog turn.
For example in the utterance transcription \textit{``Yes, the one which price is less than 50 Euros per night''}, \textit{the one} is a mention of an hotel previously introduced in the dialog.
Statistics on the corpus MEDIA are shown in table~\ref{tab:MEDIAStats}.

The task resulting from the corpus MEDIA can be modeled as a sequence labeling task by chunking the concepts over several words using the traditional \emph{BIO} notation \cite{Ramshaw95-BIO}.

Thanks to the characteristics of these two corpora, together with their relatively small size which allows training models in a reasonable time, these two tasks provide ideal settings for the evaluation of models for sequence labeling.
A comparative example of annotation, showing also the word classes available for the two tasks and mentioned above, is shown in the table~\ref{tab:ATIS-MEDIA-exemple}.

\begin{table}[t]
	    \centering
	    \scriptsize
	    \begin{tabular}{|ccc|ccc|}
	    \hline
	    \multicolumn{3}{|c|}{MEDIA} & \multicolumn{3}{|c|}{ATIS} \\
	    %\hline
	    \textbf{Words} & \textbf{Classes} & \textbf{Labels} & \textbf{Words} & \textbf{Classes} & \textbf{Labels} \\
	    \hline
        Oui & - & Answer-B & i'd & - & O \\
        l' & - & BDObject-B & like & - & O \\
        hotel & - & BDObject-I & to & - & O \\
        le & - & Object-B & fly & - & O \\
        prix & - & Object-I & Delta & airline & airline-name\\
        à & - & Comp.-payment-B & between & - & O \\
        moins & relative & Comp.-payment-I & Boston & city & fromloc.city-name\\
        cinquante & tens & Paym.-amount-B & and & - & O \\
        cinq & units & Paym.-amount-I & Chicago & city & toloc.city-name\\
        euros & currency & Paym.-currency-B & & &\\
        \hline
	    \end{tabular}
	    \caption{An example of annotated utterance transcription taken from MEDIA (left) and ATIS (right). The translation in French is \textit{``Yes, the one which price is less than 50 Euros per night''}}
	    \label{tab:ATIS-MEDIA-exemple}
    \end{table}

\begin{table}[t]
\begin{minipage}{1.0\linewidth}
    \centering
    \scriptsize
    \begin{tabular}{|l|rr|rr|rr|}
      \hline
      & \multicolumn{2}{|c|}{Training} & \multicolumn{2}{|c|}{Dev.} & \multicolumn{2}{|c|}{Test}\\
      \hline
      \# Sentences     &\multicolumn{2}{|c|}{12,908} &\multicolumn{2}{|c|}{1,259}&\multicolumn{2}{|c|}{3,005} \\
      \hline
      \hline
      & \multicolumn{1}{|c}{words} & \multicolumn{1}{c|}{concepts} &  \multicolumn{1}{|c}{words} & \multicolumn{1}{c|}{concepts} &
      \multicolumn{1}{|c}{words} & \multicolumn{1}{c|}{concepts} \\
      \hline
      \# mots          & 94,466 & 43,078 & 10,849 & 4,705 & 25,606 & 11,383 \\
      \# vocab.         &  2,210 &     99 &    838 &    66 &  1,276 &     78 \\
%      \# singletons      &    798 &     16 &    338 &     4 &    494 &     10 \\                                                                                                 
      \# OOV\%   & --     & --     &  1.33  & 0.02  &  1.39  &  0.04  \\
      \hline
    \end{tabular}
    \caption{Statistic of the corpus MEDIA}
  \label{tab:MEDIAStats}
  \end{minipage}
\end{table}

\subsection{Settings}
\label{subsec:Settings}

The RNN variant \emph{LD-RNN} has been implemented in \textit{Octave}\footnote{https://www.gnu.org/software/octave/; \\
Our code is described at http://www.marcodinarelli.it/software.php and available upon request.} using \textit{OpenBLAS} for low-level computations\footnote{http://www.openblas.net; This library allows a speed-up of roughly $330\times$ on a single matrix-matrix multiplication using $16$ cores. This is very attractive with respect to the speed-up of $380\times$ that can be reached with a GPU, keeping into account that both Octave and OpenBLAS are available for free.}.

\emph{LD-RNN} models are trained with the following procedure:

\begin{itemize}
\item Neural Network Language Models (NNLM), like the one described in \cite{Bengio03aneural}, are trained for words and labels to generate the embeddings (separately).
\item Forward and backward models are trained using the word and label embeddings trained at previous step.
\item The bidirectional model is trained using as starting point the forward and backward models trained at previous step.
\end{itemize}

We ran also some experiments using embeddings trained with \textit{word2vec} \cite{Word2Vec_Mikolov-2013}.
The results obtained are not significantly different from those obtained following the procedure described above.
This outcome is similar to the one obtained in \cite{Vukotic.etal_2015}. Since the tasks addressed in this paper are made of small data, we believe that any embedding is equally effective. In particular tools like \textit{word2vec} are designed to work on relatively big amount of data.
Results obtained with \textit{word2vec} embeddings will not be described in the following sections.

% Nombre d'époques
We roughly tuned the number of learning epochs for each model on the development data of the addressed tasks: 
$30$ epochs are used to train word embeddings, $20$ for label embeddings, $30$ for the forward and backward models, $8$ for the bidirectional model (the optimum of this model is often reached at the first epoch on the ATIS task, between the 3rd and the 5th epoch on MEDIA).
At the end of the training phase, we keep the model giving the best prediction accuracy on the development data.
We stop training the model if the accuracy is not improved for $5$ consecutive epochs (also known as \textit{Early stopping} strategy \cite{PracticalRecommendations-Bengio-2012}).

% Initialisation des poids
We initialize all the weights with the ``so called'' \textit{Xavier initialization} \cite{PracticalRecommendations-Bengio-2012}, theoretically motivated in \cite{LeakyReLU-PReLU-2015} as keeping the standard deviation of the weights during the training phase when using \emph{ReLU}, which is the type of hidden layer unit we chose for our variant of RNN.

% Hyper-paramètres (LR, LR-decay, L2, dropout)
We also tuned some of the hyper-parameters on the development data:
we found out that the best initial learning rate is $0.5$, this is linearly decreased with a value computed as the ratio between the initial learning rate and the number of epochs (\textit{Learing Rate decay}).
We combine \textit{dropout} and $L_2$ regularization \cite{PracticalRecommendations-Bengio-2012}, 
the best value for the dropout probability is $0.5$ at the hidden layer, $0.2$ at the embedding layer on ATIS, $0.15$ on MEDIA. The best coefficient ($\lambda$) for the $L_2$ regularization is $0.01$ for all the models, except for the bidirectional model where the best is $3e^{-4}$.

% Taille des couches
We ran also some experiments for optimizing the size of the different layers.
In order to minimize the time and the number of experiments, this optimization has been based on the result provided by the forward model on the two tasks, and using only words and labels as input (without word classes and character convolution, which were optimized separately).
The best size for the embeddings and the hidden layer is $200$ for both tasks.
The best size for the character convolution layer is $50$ on ATIS, $80$ on MEDIA.
In both cases, the best size for the convolution window is $1$, meaning that characters are used individually as input to the convolution.
A window of size $3$ (one character on the left, one on the right, plus the current character) gives roughly the same results, we thus prefer the simpler model.
With a window of size $5$, results starts to slightly deteriorate.

We also optimized the size of the word and label context used in the \emph{LD-RNN} variant.
On ATIS the best word context size is $11$ ($5$ on the lest, $5$ on the right plus the current word), the best label context size is $5$. On MEDIA the best sizes are $7$ and $5$ respectively.
These values are the same found in \cite{Vukotic.etal_2015} and comparable to those of \cite{Mesnil-RNN-2015}.

The best parameters found in this phase has been used to obtain \textit{baseline} models.
The goal was to understand the behavior of the models with the different level of information used: 
the word classes available for the tasks, and the character level convolution.
Some parameters needed to be re-tuned, as we will describe later on.

% Temps d'apprentissage
Concerning training and testing time of our models, 
the overall time to train and test \textit{forward}, \textit{backward} and bidirectional models, 
using only words and classes as input, is roughly $1$ hour $10$ minutes on MEDIA, $40$ minutes on ATIS.
These times go to $2$ hours for MEDIA and $2$ hours $10$ minutes for ATIS, using also word classes and character convolution as input.
All these times are measured on a \emph{Intel Xeon E5-2620} at $2.1$ \emph{GHz}, using $16$ cores.
\vspace{-0.8em}

\subsection{Results}
\label{subsec:Results}

All the results shown in this section are averages over $6$ runs.
Embeddings were learned once for all experiments.

\subsubsection{Incremental Results with Different Level of Information}

In this section we describe results obtained with incremental levels of information given as input to the models:
i) Only words (previous labels are always given as input), indicated with \emph{Words} in the tables; ii) words and classes \emph{Words+Classes}; iii) words and character convolution \emph{Words+CC}; iv) All possible inputs \emph{Words+Classes+CC}.

The results obtained on the ATIS task are shown in the table~\ref{tab:SLUATIS-Incremental}, 
results on MEDIA are in table~\ref{tab:SLUMEDIA-Incremental}.

\begin{table}[!]
    \centering
    \scriptsize
    \begin{tabular}{|l|r|r|r|}
      \hline
      Model & \multicolumn{3}{c|}{F1 measure} \\
        \hline
        \hline
								&	\textit{forward}	&	\textit{backward}		&	bidirectional \\
	\hline
	\emph{LD-RNN}	Words			&	94.23\%		&	94.30\%			&	94.45\% \\
	\emph{LD-RNN}	Words+CC			&	94.56\%		&	94.69\%			&	94.79\% \\
	\emph{LD-RNN}	Words+Classes		&	95.31\%		&	95.42\%			&	95.53\% \\
	\emph{LD-RNN}	Words+Classes+CC	&	\textbf{95.55\%}	&	\textbf{95.45\%}		&	\textbf{95.65\%} \\
      \hline
    \end{tabular}
    \caption{Results in terms of F1 measure on ATIS, using different level of information as input.}
  \label{tab:SLUATIS-Incremental}
\vspace{-1.8em}
\end{table}

\begin{table}[!]
    \centering
    \scriptsize
    \begin{tabular}{|l|r|r|r|}
      \hline
      Model & \multicolumn{3}{c|}{F1 measure} \\
        \hline
        \hline
								&	\textit{forward}	&	\textit{backward}		&	bidirectional \\
	\hline
	\emph{LD-RNN}	Words			&	85.39\%		&	86.54\%			&	87.05\% \\
	\emph{LD-RNN}	Words+CC			&	85.41\%		&	86.48\%			&	86.98\% \\
	\emph{LD-RNN}	Words+Classes		&	\textbf{85.46\%}	&	86.59\%			&	87.16\% \\
	\emph{LD-RNN}	Words+Classes+CC	&	85.38\%		&	\textbf{86.79\%}		&	\textbf{87.22\%} \\
      \hline
    \end{tabular}
    \caption{Results in terms of F1 measure on MEDIA, using different level of information as input.}
  \label{tab:SLUMEDIA-Incremental}
\vspace{-1.8em}
\end{table}

Results in these tables show that models have a similar behavior on the two tasks.
In particular on ATIS, adding the different level of information results improve progressively and the best performance is obtained integrating words, labels and character convolution, 
though some of the improvements do not seem statistically significant, taking into account the small size of this corpus.

This observation is confirmed by results obtained on MEDIA, where adding the character level convolution leads to a slight degradation of performances.
In order to understand the reason of this behavior we analyzed the training phase on the two tasks.
We found out that the main problem was an hidden layer saturation: with the number of hidden neurons chosen in the preliminary optimization phase using only words (and labels), the hidden layer was not able to model the whole information richness provided by all the inputs at the same time.
We ran thus some experiments using a larger hidden layer with size $256$, 
which gave the results shown in the two tables with the model \emph{LD-RNN Words+Classes+CC}.
For lack of time we did not further optimized the size of the hidden layer.

Beyond all of that, results shown in the table~\ref{tab:SLUATIS-Incremental} and \ref{tab:SLUMEDIA-Incremental} are very competitive, as we will discuss in the next section.

\subsubsection{Comparison with the State-of-the-Art}
\label{subsubsec:Comparison}

In this section we compare our results with the best results found in the literature.
In order to be fair, the comparison is made using the same input information: words and classes.
In the tables we use \emph{E-RNN} for Elman RNN, \emph{J-RNN} for Jordan RNN, \emph{I-RNN} for the improved RNN proposed by \cite{DinarelliTellier-RNN-TALN2016}.\footnote{This is a publication in French, but results in the tables are easy to understand and directly comparable to our results.}

In order to give an idea of how our RNN variant compares to \textit{LSTM+CRF} models like the one of \cite{Ma-Hovy-ACL-2016}, we ran an experiment on the Penn Treebank \cite{Marcus93buildinga}.
With a similar data pre-processing, exactly the same data split, using a \textit{sigmoid} activation function, and using only words as input, the \emph{LD-RNN} variant achieves an accuracy of $96.83$.
This is comparable to the $96.9$ achieved by the \textit{LSTM+CRF} model of \cite{Ma-Hovy-ACL-2016} without pre-trained embeddings.\footnote{We did not run further experiments because without a GPU, experiments on the Penn Treebank are still quite expensive.}

Results on the ATIS task are shown in table~\ref{tab:SLUATIS}.
On this task we compare to results published in \cite{Vukotic.etal_2016} and \cite{DinarelliTellier-RNN-TALN2016}.

The results in the table~\ref{tab:SLUATIS} show that all models obtain a good performance on this task, 
always higher than $94.5$ $F1$. This confirm what we anticipated in the previous section concerning how easy is this task.

The GRU RNNs of \cite{Vukotic.etal_2016} and our variant \emph{LD-RNN} obtain equivalent results ($95.53$), which is slightly better than all the other models, in particular with the bidirectional models.
This is a good outcome, as our variant of RNN obtains the same result as GRU while using much less parameters (see section~\ref{subsec:Complexity} for RNNs complexity).
Indeed LSTM and GRU are considered very effective models for learning very long contexts.
The way they are used in \cite{Vukotic.etal_2016} allows to learn long contexts on the input side (words), they are not adapted however to learn also long label contexts, which is what we do in this paper with our variant.
The fact that the best word context on this task is made of $11$ words, show that this is the most important information to obtain good results on this task. It is thus not surprising that the GRU RNN achieves such good performance.

Comparing our results on the ATIS task with those published in \cite{DinarelliTellier-RNN-TALN2016} with a Jordan RNN, which uses the same label context as our models, we can conclude that the advantage in the variant \emph{LD-RNN} is given by the use of label embeddings and their combination at the hidden layer.

\begin{table}[!]
    \centering
    \scriptsize
    \begin{tabular}{|l|r|r|r|}
      \hline
      Model & \multicolumn{3}{c|}{F1 measure} \\
        \hline
        \hline
										&	\textit{forward}	&	\textit{backward}	&	bidirectional \\
	\hline
	\cite{Vukotic.etal_2016} LSTM				&	95.12\% 		& 	-- 			&	95.23\% \\
	\cite{Vukotic.etal_2016} GRU				&	\textbf{95.43\%} 		& 	-- 			&	\textbf{95.53\%} \\
	\hline
	\cite{DinarelliTellier-RNN-TALN2016} E-RNN	&	94.73\%		&	93.61\%		&	94.71\% \\
	\cite{DinarelliTellier-RNN-TALN2016} J-RNN		&	94.94\%		&	94.80\%		&	94.89\% \\
	\cite{DinarelliTellier-RNN-TALN2016} I-RNN		&	95.21\%		&	94.64\%		&	94.75\% \\
	\hline
	\emph{LD-RNN}	Words+Classes				&	95.31\%		&	95.42\%		&	\textbf{95.53\%} \\
      \hline
    \end{tabular}
    \caption{Comparison of our results on the ATIS task with the literature, in terms of F1 measure.}
  \label{tab:SLUATIS}
\vspace{-1.8em}
\end{table}

\begin{table}[!]
    \centering
    \scriptsize
    \begin{tabular}{|l|r|r|r|}
      \hline
      Model & \multicolumn{3}{c|}{F1 measure} \\
        \hline
        \hline
										&	\textit{forward}	&	\textit{backward}	&	bidirectional \\
	\hline
	\cite{Vukotic.etal_2015} CRF				&	\multicolumn{3}{c|}{86.00\%} \\
	\hline
	\cite{Vukotic.etal_2015} E-RNN				&	81.94\%		&	-- 			&	-- \\
	\cite{Vukotic.etal_2015} J-RNN				&	83.25\%		&	-- 			&	-- \\
	\hline
	\cite{Vukotic.etal_2016} LSTM				&	81.54\% 		& 	-- 			&	83.07\% \\
	\cite{Vukotic.etal_2016} GRU				&	83.18\% 		& 	-- 			&	83.63\% \\
	\hline
	\cite{DinarelliTellier-RNN-TALN2016} E-RNN	&	82.64\%		&	82.61\%		&	83.13\% \\
	\cite{DinarelliTellier-RNN-TALN2016} J-RNN		&	83.06\%		&	83.74\%		&	84.29\% \\
	\cite{DinarelliTellier-RNN-TALN2016} I-RNN		&	84.91\%		&	86.28\%		&	86.71\% \\
	\hline
	\emph{LD-RNN}	Words+Classes				&	\textbf{85.46\%}		&	\textbf{86.59\%}		&	\textbf{87.16\%} \\
      \hline
    \end{tabular}
    \caption{Comparison of our results on the MEDIA task with the literature, in terms of F1 measure.}
  \label{tab:SLUMEDIA}
\vspace{-1.8em}
\end{table}

This conclusion is more evident if we compare results obtained with RNNs using label embeddings with the other RNNs on the MEDIA task.
This comparison is shown in table~\ref{tab:SLUMEDIA}.
As we mentioned in the section~\ref{subsec:Corpora}, this task is very challenging for several reason, 
but in the context of this paper we focus on the label dependencies that we claim we can effectively model with our RNN variant.

In this context we note that a traditional Jordan RNN, the \emph{J-RNN} of \cite{DinarelliTellier-RNN-TALN2016}, 
which is the only traditional model to explicitly use previous label information as context, is more effective than the other traditional models, including LSTM and GRU ($84.29$ F1 with J-RNN, $83.63$ with GRU, second best model among traditional RNNs).
We note also that on MEDIA, CRFs, which are models specifically designed for sequence labeling, are by far more effective than the traditional RNNs ($86.00$ F1 with the CRF of \cite{Vukotic.etal_2015}).

The only models outperforming CRFs on the MEDIA task are the \emph{I-RNN} model of \cite{DinarelliTellier-RNN-TALN2016} and our \emph{LD-RNN} variant, both using label embeddings.

\begin{table}[!]
    \centering
    \scriptsize
    \begin{tabular}{|l|r|}
      \hline
      Model & CER \\
        \hline
        \cite{Dinarelli.etAl-SLU-RR-2011} CRF			&	11.7\% \\
        \cite{dinarelli2011-emnlp} CRF				&	11.5\% \\
        \cite{Hahn.etAL-SLUJournal-2010} CRF		&	10.6\% \\
        \hline
	\emph{LD-RNN}	Words					&	10.73\% (10.63) \\
	\emph{LD-RNN}	Words+Classes				&	10.52\% (10.15) \\
	\emph{LD-RNN}	Words+Classes+CC			&	\textbf{10.41\% (10.09)} \\
      \hline
    \end{tabular}
    \caption{\small{Results on the MEDIA task in terms of \emph{Concept Error Rate} (CER), compared with the best results published so far on this task.}}
  \label{tab:SLUMEDIA-CER}
\vspace{-1.8em}
\end{table}

Even if results on MEDIA discussed so far are very competitive, 
this task has been designed for Spoken Language Understanding (SLU) \cite{demori08-SPM}.
In SLU the goal is to extract a correct semantic representation of a sentence, allowing a correct interpretation of the user will by the spoken dialog system.
While the F1 measure is strongly correlated with SLU evaluation metrics, the evaluation measure used most often in the literature is the \emph{Concept Error Rate} (CER). CER is defined exactly in the same way as \emph{Word Error rate} in automatic speech recognition, where words are replaced by concepts.\footnote{The errors made by the system are classified as Insertions (I), Deletions (D) and Substitutions (S). The sum of these errors is divided by the number of concepts in the reference annotation (R): $CER = \frac{I + D + S}{R}$.}

In order to place our results on an absolute ranking among models designed for the MEDIA task, 
we propose a comparison in terms of CER to the best models published in the literature, 
namely \cite{Hahn.etAL-SLUJournal-2010}, \cite{dinarelli2011-emnlp} and \cite{Dinarelli.etAl-SLU-RR-2011}.
This comparison is shown in table~\ref{tab:SLUMEDIA-CER}.

The best individual models published by \cite{Hahn.etAL-SLUJournal-2010}, \cite{dinarelli2011-emnlp} and \cite{Dinarelli.etAl-SLU-RR-2011} are CRFs, achieving a CER of $10.6$, $11.5$ and $11.7$, respectively.
These models use both word and classes, and a rich set of lexical features such like word prefixes, suffixes, word capitalization information etc.
We note that the large gap between these CRF models is due to the fact that the CRF of \cite{Hahn.etAL-SLUJournal-2010} is trained with an improved margin criterion, similar to the large margin principle of \emph{SVM} \cite{Herbrich.etAL-LargeMargin-2000,hahn09-Interspeech}.
We note also that comparing significance tests published in \cite{Dinarelli.etAl-SLU-RR-2011}, a difference of $0.1$ in CER is already statistically significant. Since results in this paper are higher, we hypothesize than even smaller gains are significant.

Our best \emph{LD-RNN} model achieve a CER of $10.41$.
To the best of our knowledge this is the best CER obtained on the MEDIA task with an individual model.
Moreover, instead of taking the mean of CER of several experiments, following a strategy similar to \cite{RNNforLU-Interspeech-2013}, one can run several experiments and keep the model obtaining the best CER on the development data of the target task.
Results obtained using this strategy are shown in table~\ref{tab:SLUMEDIA-CER} between parenthesis.
The best result obtained by our \emph{LD-RNN} is a CER of $10.09$, the best absolute result on this task so far, even better than the \emph{ROVER} model \cite{fiscus97-rover} used in \cite{Hahn.etAL-SLUJournal-2010}, 
which combines $6$ individual models, including the individual CRF model achieving $10.6$ CER.
\vspace{-0.8em}

\subsection{Results Discussion}
\label{subsec:Analyses}

In order to understand the high performances of the \emph{LD-RNN} variant on the MEDIA task, 
we made some simple analyses on the model output, 
comparing them to the output of a Jordan RNN trained with our own system in the same conditions as \emph{LD-RNN} models.
The main difference between these two models is the general tendency of the Jordan RNN to split a single concept into two or more concepts, mainly for concepts instantiated by long surface forms, such like \textit{command-tache}.
This concept is used to mean the general user will in a dialog turn (e.g. \textit{Hotel reservation}, \textit{Price information} etc.). The Jordan RNN often split this concept into several concepts by introducing a void label, associated to a stop-word.
This is due to the limitation of this model to take relatively long label context into account, even if it is the only traditional RNN using explicitly previous labels as context information.

Surprisingly, \emph{LD-RNN} never makes this mistake and in general never makes segmentation errors (concerning the BIO formalism). This can be due to two reasons. The first is that label embeddings learns similar representations for semantically similar labels. This allows the model to correctly predict start-of-concept (B) even if the target word has been seen in the training set only as continuation-of-concept (I), or viceversa, as the two labels acquire very similar representations. The second reason, which is not in mutual exclusion with the first, is that the model factorizes information acquired on similar words seen associated to start-of-concept labels. Thus if a word has not been seen associated to start-of-concept labels, but similar words do, the model is still able to provide the correct annotation. This second reason is what made neural networks popular for learning word embeddings in earlier publications \cite{Bengio03aneural}.
In any case, in our experience, we never observed such precise behavior even with CRF models tuned for the MEDIA task.
For this reason we believe \emph{LD-RNN} deserves the name of \emph{Label Dependencies aware} RNN.

Still \emph{LD-RNN} makes mistakes, which means that once a label annotation starts for a target word, even if the label is not the correct one, the same label is kept even if the following words provide evidence that the correct label is another one. \emph{LD-RNN} tends to be coherent with previous labeling decisions. This behavior is due to the use of a local decision function which definitely relies heavily on the label embedding context, but it doesn't prevent the model from being very effective.
Interestingly, this behavior suggests that \emph{LD-RNN} could still benefit from a CRF neural layer like those used in \cite{huang2015bidirectional,lample2016neural,Ma-Hovy-ACL-2016}. We leave this as future work.

\vspace{-0.8em}

\section{Conclusion}
\label{sec:Conclusions}

In this paper we proposed a new variant of RNN for sequence labeling using a wide context of label embeddings in addition to the word context to predict the next label in a sequence.
We motivated our variant as being more effective at modeling label dependencies.
Results on two Spoken Language Understanding tasks show that i) on a simple task like ATIS our variant achieves the same performance as much more complex models such as LSTM and GRU, which are claimed the most effective RNNs; ii) on the MEDIA task, where modeling label dependencies is crucial, our variant outperforms by a large margin all the other RNNs, including LSTM and GRU.
When compared to the best models of the literature in terms of Concept Error Rate (CER), 
our RNN variant results to be more effective, achieving a state-of-the-art CER of $10.09$.

\section{Acknowledgements}

This work has been partially funded by the French ANR project Democrat ANR-15-CE38-0008.

%% THEEND

\bibliographystyle{splncs}
\bibliography{2016_arXiv_NewRNN_author-final}
\end{document}